\documentclass[conference]{IEEEtran}
\IEEEoverridecommandlockouts

\usepackage{cite}
\usepackage{amsmath,amssymb,amsfonts}
\usepackage{algorithmic}
\usepackage{graphicx}
\usepackage{textcomp}
\usepackage{xcolor}
\usepackage{float}
\usepackage{subfigure}
\usepackage{lipsum}
\usepackage{hyperref}
\usepackage{booktabs}
\newcommand{\greenbf}[1]{\textbf{\textcolor{green!60!black}{#1}}}
\def\BibTeX{{\rm B\kern-.05em{\sc i\kern-.025em b}\kern-.08em
    T\kern-.1667em\lower.7ex\hbox{E}\kern-.125emX}}
\begin{document}

\title{GHaLIB: A Multilingual Framework for Hope Speech Detection in Low-Resource Languages}

\author{
    \IEEEauthorblockN{Ahmed Abdullah}
    \IEEEauthorblockA{\textit{School of Computing} \\
    \textit{FAST-National University}\\
    Lahore, Pakistan \\
    l227503@lhr.nu.edu.pk}
    \and
    \IEEEauthorblockN{Haroon Mahmood}
    \IEEEauthorblockA{\textit{College of Engineering} \\
    \textit{Al Ain University}\\
    Al-Ain, UAE \\
    haroon.mahmood@aau.ac.ae}
    \and
    \IEEEauthorblockN{Sana Fatima}
    \IEEEauthorblockA{\textit{School of Computing} \\
    \textit{FAST-National University}\\
    Lahore, Pakistan \\
    sana.fatima@nu.edu.pk}
}

\maketitle

\begin{abstract}
Hope speech has been relatively underrepresented in Natural Language Processing (NLP). Current studies are largely focused on English, which has resulted in a lack of resources for low-resource languages such as Urdu. As a result, the creation of tools that facilitate positive online communication remains limited. Although transformer-based architectures have proven to be effective in detecting hate and offensive speech, little has been done to apply them to hope speech or, more generally, to test them across a variety of linguistic settings. This paper presents a multilingual framework for hope speech detection with a focus on Urdu. Using pretrained transformer models such as XLM-RoBERTa, mBERT, EuroBERT, and UrduBERT, we apply simple preprocessing and train classifiers for improved results. Evaluations on the PolyHope-M 2025 benchmark demonstrate strong performance, achieving F1-scores of 95.2\% for Urdu binary classification and 65.2\% for Urdu multi-class classification, with similarly competitive results in Spanish, German, and English. These results highlight the possibility of implementing existing multilingual models in low-resource environments, thus making it easier to identify hope speech and helping to build a more constructive digital discourse.
\end{abstract}

\begin{IEEEkeywords}
Natural language processing, Hope speech detection, Low-resource languages, Urdu language, Multilingual transformer models, Text classification, Support Vector Machine (SVM), LightGBM
\end{IEEEkeywords}

\section{Introduction}
Understanding public sentiment across multiple domains, such as social media, news, and customer feedback, is an important activity in sentiment analysis within Natural Language Processing (NLP). However, analyzing sentiments that involve sarcasm and hope speech presents a different type of challenge due to the mixed nature of natural language. Sarcasm is particularly elusive, as it involves utterances that convey the contrary of what is literally meant \cite{codabench_garcia2023lgtb}. Most sentiment analysis models crumble when tackling sarcasm and false hope; they frequently misclassify the underlying sarcasm as emotions that are neutral or positive, while the statement actually conveys a very negative connotation \cite{Du2022}.

\begin{figure}[h!]
    \centering
    \includegraphics[width=0.5\textwidth]{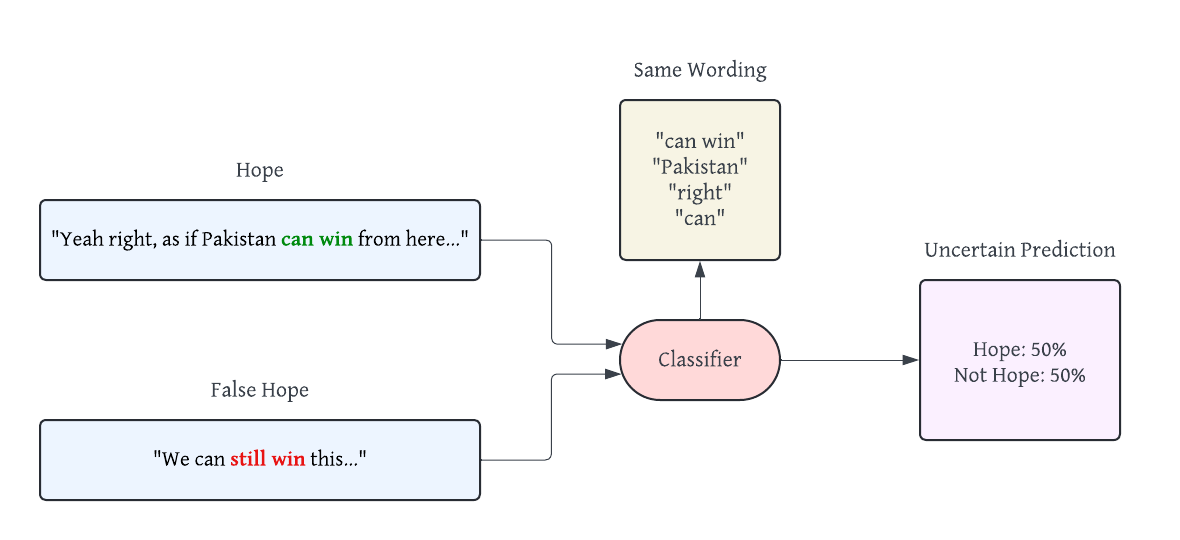}
    \caption{Uncertain prediction caused by shared wording in hope sentiments.}
    \label{fig:hope_speech_problem}
\end{figure}

As in the case of the examples in Figure~\ref{fig:hope_speech_problem}, "can win" and "still win," expressions may be conveyed in a contrary manner within any given context. Hence, the author's intentions may often not be detected from surface features of the text. The same words may give encouragement in one instance and sarcasm or hopelessness in another, making it difficult for a classifier to determine whether a piece of text truly expresses hope or the exact opposite \cite{codabench_garcia2024iberlef}\cite{codabench_balouchzahi2025polyhopem}. Therefore, the classification usually leads to greater uncertainties and misclassifications, which emphasizes the need for deeper contextual and pragmatic understanding in automated systems.

The following categories of Hope Speech are generally used in the literature:

\begin{itemize}
    \item \textbf{Generalized Hope:} A belief that things will get better in future.
    \\ Example: \textit{"Things will improve over time."}
    This type of hope reflects a belief that things will improve, without pointing to a specific action or result. It is often expressed in uncertain situations where the future is unpredictable and serves to comfort others \cite{codabench_balouchzahi2023polyhope}. 

    \item \textbf{Realistic Hope:} Hope for something possible to achieve. 
    \\ Example: \textit{"I can pass the exam if I work hard for it." }
    Realistic hope is based on goals that are possible to reach. It connects optimism with effort or conditions that make success likely, making it practical and useful in real situations \cite{codabench_balouchzahi2023polyhope}.

    \item \textbf{Unrealistic Hope:} Hope for something unlikely or impossible to achieve. 
    \\ Example: \textit{"If I jump high enough, I can reach the moon."}
    This type of hope focuses on outcomes that cannot realistically happen. While it may still give emotional relief, it often creates false expectations \cite{codabench_balouchzahi2023polyhope}.

    \item \textbf{Not Hope:} Texts that do not show hope. 
    \\ Example: \textit{"Nothing is going to change."}
    This category includes statements that lack optimism and instead show negativity, doubt, or resignation. It is important for separating hopeful from non-hopeful content in analysis. It can also refer to as the text that not even related to hopelessness context \cite{codabench_balouchzahi2023polyhope}. 
\end{itemize}

The main contributions of this work are:  
\begin{itemize}  
    \item We propose a multilingual framework for hope speech detection with a focus on Urdu and English.  
     \item We evaluate multiple transformer models and machine learning classifiers on the PolyHope-M 2025 dataset.  
    \item We achieve strong results for Urdu and English, and show competitive performance in other languages.
    \item We release our code and resources at \href{https://github.com/ahmedembeddedxx/GHaLIB}{GHaLIB}.  
\end{itemize}  

The rest of the paper is organized as follows. 
Section~\ref{sec:related} reviews related work and literature. 
Section~\ref{sec:data} presents exploratory data analysis, highlighting how different features can contribute to the task. 
Section~\ref{sec:method} describes our methodology. 
Section~\ref{sec:results} reports experiments and results. 
Section~\ref{sec:future} discusses future work and provides the conclusion.

\section{Literature Review} \label{sec:related}
\subsection{Early Feature-Based Approaches to Hope Speech Detection}
Lexicon-based approaches have been future-oriented in speech evaluation with a focus on the initial development of the lexicon and have relied heavily on pre-defined lists of positive terms. The limitations of hard-coded vocabulary included failing to perceive the language's contextual sense, open-world benchmarks and the various nuances, which led the lexicon-based approaches lead to very high false positive incidents \cite{zechner2013textclassification}. Palakodety \textit{et al.} (2019) \cite{palakodety2019hopespeech} suggested one of the first computational studies in detecting hope speech, focusing on comments on Hindi and English YouTube channels in response to the 2019 India-Pakistan crisis. The constructed hope speech corpus was trained with a logistic regression classifier using L2-regularization by the authors, having n-gram features along with another linguistic indicator. Palakodety ran experiments using an 80/10/10 split for training, validation, and testing, and averaged the results over 100 random splits to make sure the results were reliable. The configuration that worked ideally combined n-grams, indicator features, and FastText embeddings, leading to an F1-score of 78.5\%, precision of 82.0\%, recall of 75.3\%, and an AUC of 95.5\% on self-curated benchmark. These results overlapped an arbitrary standard for a general-purpose sentiment analysis system. These results significantly outperformed a general-purpose sentiment analysis baseline  of Stanford CoreNLP, highlighting that hope speech detection is a distinct challenge requiring dedicated methods. Their work established early benchmarks and demonstrated that targeted feature engineering combined with classical machine learning can bring strong results on the hope classification tasks.

\subsection{Traditional Machine Learning Models}
With the emergence of traditional Machine Learning (ML), researchers began employing algorithms such as Support Vector Machines (SVM), Naive Bayes, and logistic regression for hope speech detection. Balouchzahi \textit{et al.} (2023) \cite{codabench_balouchzahi2023polyhope} implemented a logistic regression model for Urdu hope speech classification, attaining an F1-score of 75.9\% in binary classification and a F1 (macro) score of 48.0\% in multiclass settings. Similar to this, Sidorov \textit{et al.} (2024) \cite{sidorov2024mindhope} demonstrated the use of various ML techniques, in addition to deep learning, for multiclass hope speech detection in Spanish and German. Their experiments reported F1-scores of 68.0\% for Spanish and 69.8\% for German through classical machine learning classifiers, proving their viability in the task. Yet, although these are still promising results, traditional ML models in general suffered in multilingual and low-resource contexts, often relying heavily on handcrafted features and language-specific preprocessing. These limitations prompted the shift of the field toward embedding-based and deep learning architectures for more robust and scalable hope speech detection \cite{sidorov2024mindhope}.

\subsection{Deep Learning and Embedding-Based Methods}
The development of deep learning architectures using word embeddings marked a substantial advance in text classification tasks. Minaee \textit{et al.} (2021) \cite{Minaee2021} provides a highly appreciable survey of deep learning models, some 150 in total, including Convolution Neural Networks (CNNs) \cite{oshea2015introductionconvolutionalneuralnetworks}, Recurrent Neural Networks (RNNs) \cite{schmidt2019recurrentneuralnetworksrnns}, and attention-based architectures \cite{vaswani2023attentionneed}, based on their applications to recognized text classification benchmarks. Their empirical experiment has proved that quantitatively, models like CNNs, LSTMs \cite{vennerød2021longshorttermmemoryrnn}, and BiLSTMs \cite{vennerød2021longshorttermmemoryrnn} often outperform baseline models of machine learning when pretrained word-level embeddings like Word2Vec \cite{mikolov2013efficientestimationwordrepresentations} or GloVe \cite{Brochier_2019} are used. Most of the deep learning models achieved 2–7\% improvement in accuracy, depending on the task, with BiLSTM with attention being the greatest on average \cite{codabench_balouchzahi2023polyhope}. An even higher accuracy result of 96.5\% was achieved by Alshahrani \textit{et al.} (2020) \cite{alshahrani2020optimism} using an XLNet-based \cite{yang2020xlnetgeneralizedautoregressivepretraining} method on optimism classification in Twitter messages, But it wasn’t clear that CNN or LSTM models using embeddings worked well in multilingual settings. The findings by Alshahrani \textit{et al.} and Balouchzahi \textit{et al.} once again highlight the proof of purely embedding-based limitations with high computation requirements.

\subsection{Transformer-Based Models}
Models like Bidirectional Encoder Representations from Transformers (BERT) \cite{devlin2019bertpretrainingdeepbidirectional} and its versions have changed how text classification works. They have reached top performance on many tests, with high accuracy and F1-scores. However, it brought into focus problems concerning computation and biases present in the training data \cite{Ramos2024}\cite{Fields2024}. In the last few years, the transformer architecture has taken center stage in hope speech classification and has reported very good results across multiple languages tasks. In an extensive set of evaluations, Sidorov \textit{et al.} (2023) \cite{sidorov2023regret} put several transformer models (e.g., BERT, mBERT, XLM-RoBERTa) to the test over datasets for the detection of hope and regret, their best-performing transformer setup attained a F1 (macro) score of 72.0\% for hope speech, significantly outperforming classical machine learning algorithm's previous benchmark. Based on similar approach, the MIND-HOPE by Sidorov \textit{et al.} (2024) \cite{sidorov2024mindhope} introduced multiclass classification in Spanish and German. They found that monolingual BERT models (\verb |bert-base-german-dbmdz-uncased| for German and \verb |bert-base-spanish-wwm-uncased| for Spanish) outperformed multilingual models, reporting F1 (macro) scores of 69.8\% for German and 68.01\% for Spanish.

Employing two-level transformer pipelines on Urdu, German, and Spanish datasets, the PolyHope series \cite{balouchzahi2025urduhope} \cite{codabench_balouchzahi2025polyhopem}\cite{codabench_garcia2024iberlef} addressed low-resourced languages. Balouchzahi \textit{et al.} (2025) \cite{balouchzahi2025urduhope} reported an F1-score of 75.9\% for Urdu binary classification along with a macro F1 of 48.01\% on Urdu multiclass detection, whereas in the PolyHope-M shared task \cite{codabench_balouchzahi2025polyhopem}, the cross-lingual F1 (macro) scores were 83.7\% (Urdu binary), 76.2\% (Spanish binary), and 73.2\% (German binary). Khalid \textit{et al.} (2021) \cite{khalid2021rubertbilingualromanurdu} introduced RUBERT (or UrduBERT), a bilingual Roman Urdu BERT, emphasizing the importance of pretraining specific to code-mixed and low-resource hope speech contexts. Shared tasks like IberLEF \cite{codabench_garcia2024iberlef} shows that, under fine-tuning, transformer models surpass any known deep learning or classical approaches to date. In short, these results show that using transformers tailored for each language or domain is needed.

\subsection{Recent Advances and Large Language Models}
Modern research has been focusing on Large Language Models (LLMs), which have shown great results in understanding text and generating text. For example, the RoBERTa model which was specially fine-tuned for multilingual hate speech detection, achieved an impressive accuracy of 94.0\% and was complemented by an F1-score of 80.2\% in the analysis of English and Russian multilingual datasets \cite{Ramos2024}. However, as these further developments are made, concerns about algorithmic bias, hallucinations and benchmark standardization are still point of concern \cite{Malik2023}. Other LLMs, like GPT-4 and R1, are highly effective on zero-shot and few-shot classification tasks, but highly vulnerable to prompt structure \cite{Malik2023}. Chung \textit{et al.} (2022) \cite{chung2022scalinginstructionfinetunedlanguagemodels} found more reliable gains in classification tasks with instruction-tuning LLMs, including Tuner-5, whereas Puri \textit{et al.} (2019) \cite{puri2019zeroshottextclassificationgenerative} and Sun \textit{et al.} (2023) \cite{sun2023textclassificationlargelanguage} noted that prompt generation and generative LLMs still have to be highly-careful to create, yet they still face further challenges in low-resource or nuanced environments.

\section{Dataset and Exploratory Analysis} \label{sec:data}
\begin{figure}[H]
    \centering
    \includegraphics[width=1\columnwidth]{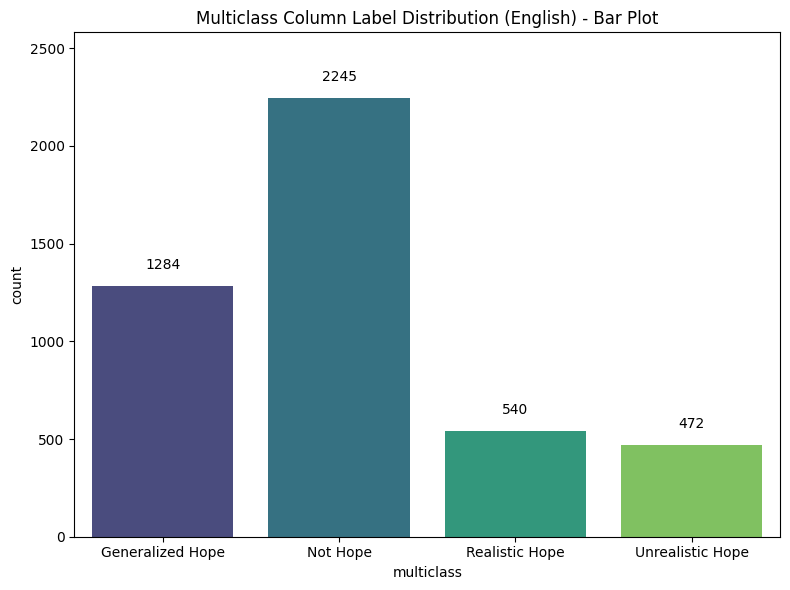}
    \caption{English dataset label distribution}
    \label{fig:eng_dist}
\end{figure}

\begin{figure}[H]
    \centering
    \includegraphics[width=1\columnwidth]{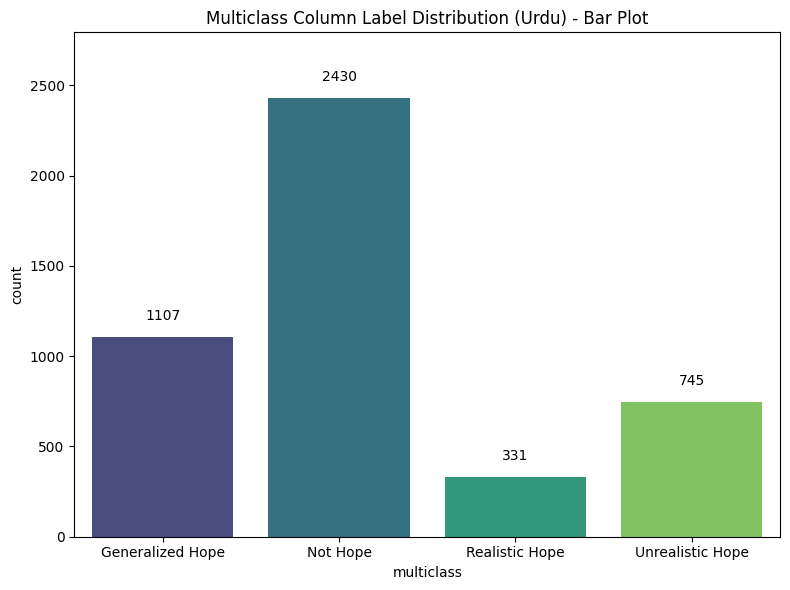}
    \caption{Urdu dataset label distribution}
    \label{fig:urdu_dist}
\end{figure}

\begin{figure}[H]
    \centering
    \includegraphics[width=1\columnwidth]{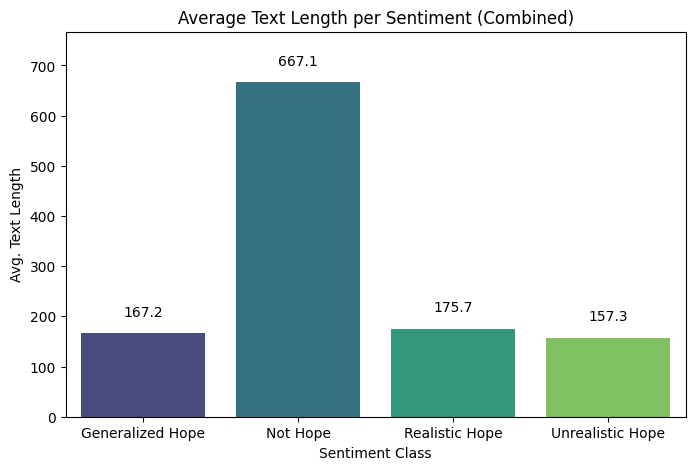}
    \caption{Average text length per sentiment category.}
    \label{fig:eng_urdu_length_per_sentiment}
\end{figure}

\begin{figure}[H]
    \centering
    \includegraphics[width=1\columnwidth]{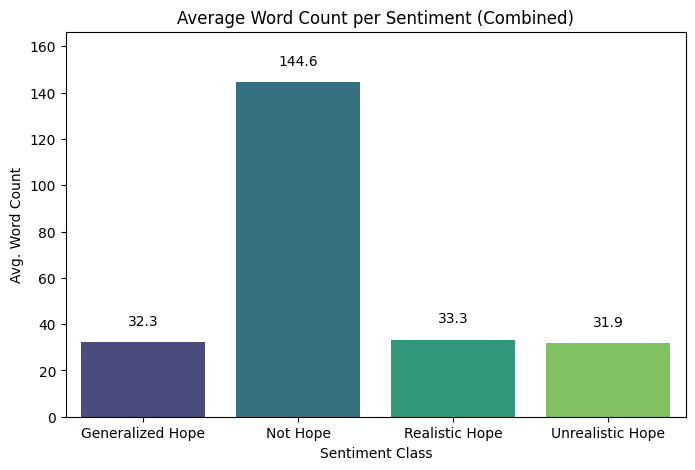}
    \caption{Average word count per sentiment category.}
    \label{fig:eng_urdu_wordcount_per_sentiment}
\end{figure}

\begin{figure}[H]
    \centering
    \includegraphics[width=0.45\textwidth]{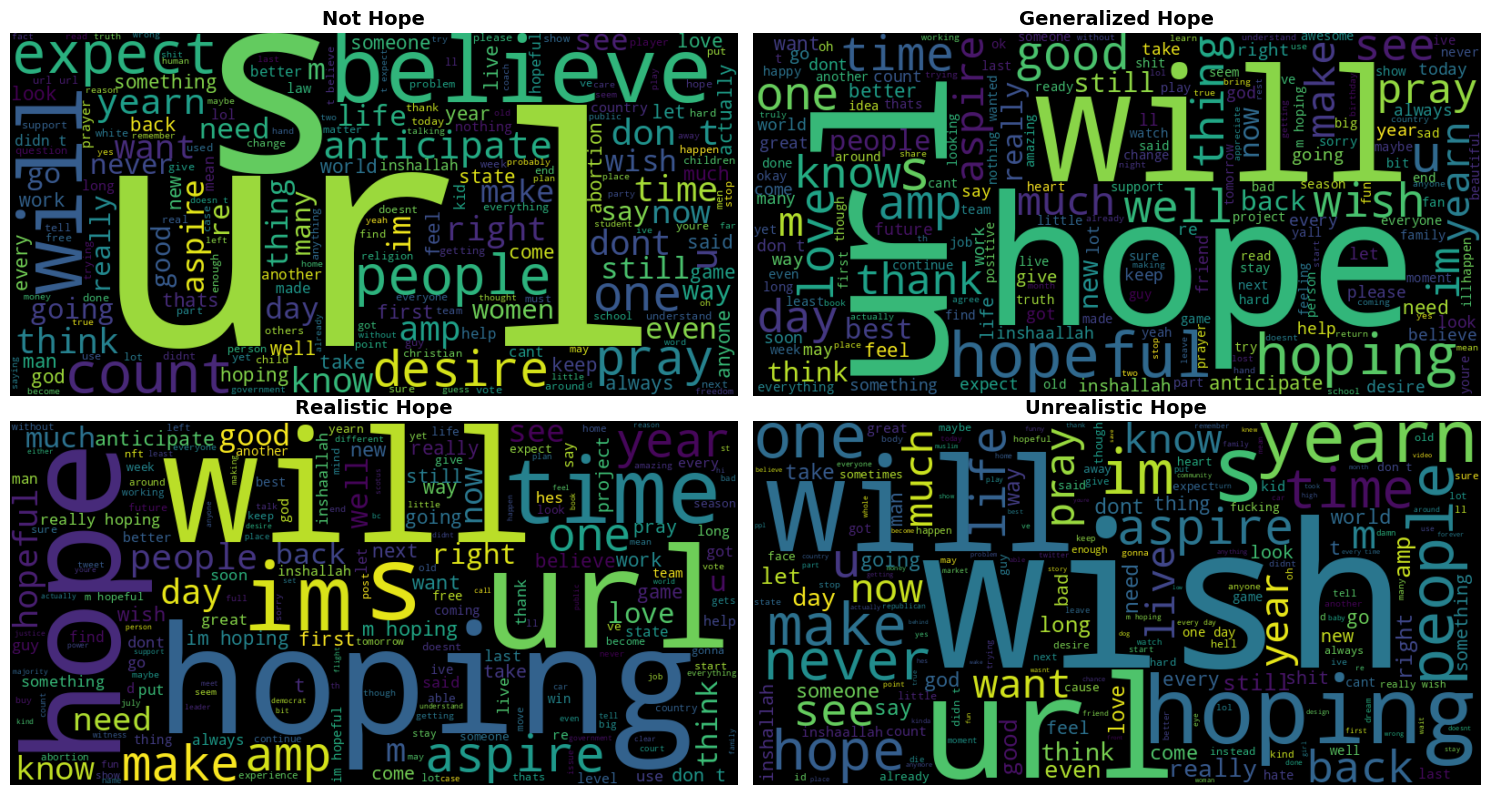}
    \caption{Word clouds for English hope speech categories showing characteristic vocabulary patterns for each sentiment class. }
    \label{fig:eng_wordclouds}
\end{figure}

We will analyze the distribution of the labels in the English and Urdu training datasets by multilingual PolyHope-M 2025 corpus \cite{codabench_balouchzahi2025polyhopem} to understand the characteristics of the datasets. Figures~\ref{fig:eng_dist} and \ref{fig:urdu_dist} show the distribution of classes for both languages, where "Not Hope" is the most populated category, followed by Generalized Hope. The information captured from the averages of text length and word count per sentiment category is presented in Figure~\ref{fig:eng_urdu_length_per_sentiment} and Figure~\ref{fig:eng_urdu_wordcount_per_sentiment}, and shows interesting patterns across the languages. Realistic Hope expressions in English tend to be longer, whereas Generalized Hope statements in Urdu are less concise. Finally, the word clouds in Figure~\ref{fig:wordclouds} and Figure~\ref{fig:eng_wordclouds} provide qualitative insight into the most frequent words used in each language in relation to their association with the Hope categories. From the English word clouds and Urdu examples, it can be seen that Realistic Hope generally involves more concrete terms, whereas Generalized Hope contains more abstract concepts.

The label distributions reflect an evident imbalance in both the English and Urdu datasets; Not Hope dominates the samples, followed by a smaller but significant portion of Generalized Hope. Realistic and Unrealistic Hope, however, are underrepresented, making it more difficult to train balanced classifiers. Text-length analysis shows interesting contrasts: in English, the expression of Realistic Hope tends to involve longer sentences and a greater number of words, whereas in Urdu, messages of Not Hope are far longer and heavier in their use of words compared to the hopeful categories. This suggests that there are stylistic and cultural distinctions in how hope or its absence is expressed in the two languages. Word clouds further substantiate this tendency, in English, positive hopeful expressions often use down-to-earth, action-oriented terminology such as will, make, and believe; in contrast, Urdu hopeful texts rely more on religious and motivational language. Our analysis and findings suggests that effective multilingual hope speech models must account for both language use and cultural context.

\begin{figure}[H]
    \centering
    \includegraphics[width=0.45\textwidth]{imgs/urdu-wordcloud.png}
    \caption{Word clouds for Urdu hope speech categories showing characteristic vocabulary patterns for each sentiment class.}
    \label{fig:wordclouds}
\end{figure}

\section{Methodology} \label{sec:method}

\begin{figure*}[!t]
    \centering
    \includegraphics[width=1\linewidth]{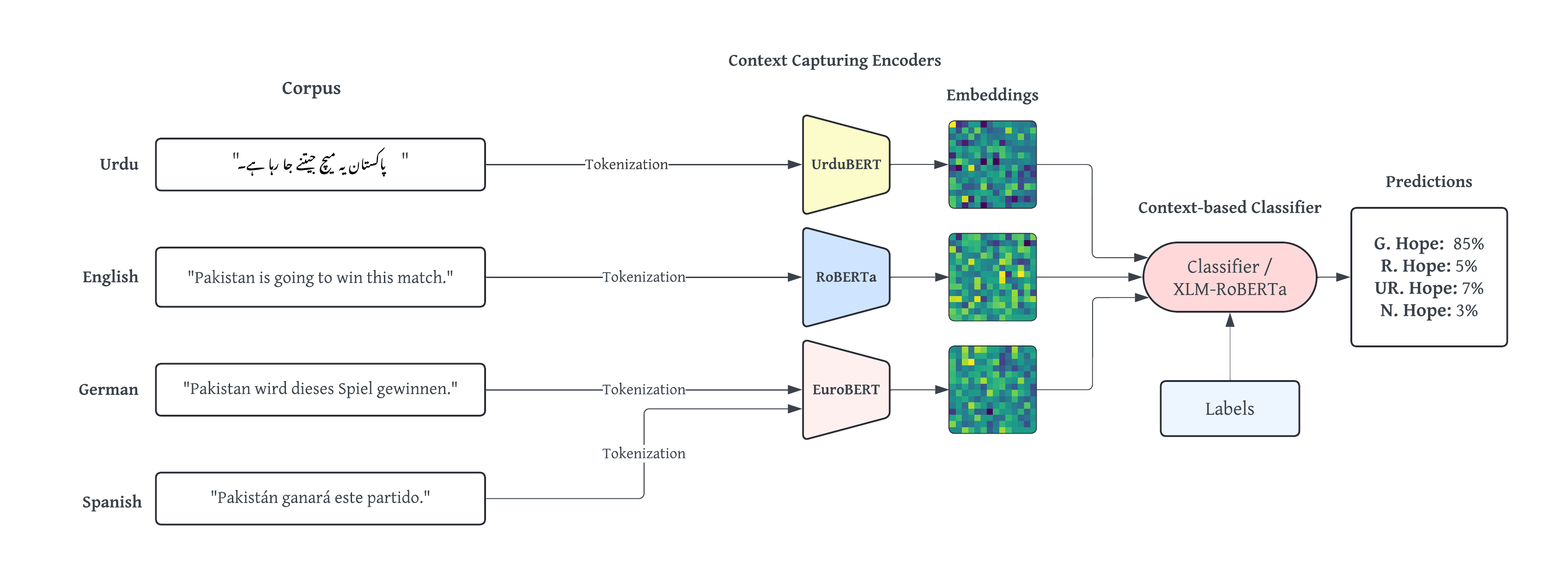}
    \caption{Proposed architecture of GHaLIB for classifying sentiment in multiple languages using XLM-RoBERTa backbone.}
    \label{fig:GHaLIB}
\end{figure*}

We present a classification system that effectively balances sensitivity and specificity. Our approach addresses common issues with cross-lingual generalization and under-represented languages by combining language-specific encoders with a multilingual backbone. This design enables our model to perform reliably across multiple languages without risking the performance. The goal, therefore, is to create a solution that scales well without losing performance across languages, while also maintaining computational efficiency for real-world deployment.

The model has been trained and evaluated on the curated multilingual PolyHope-M 2025 corpus comprising hope speech and non-hope speech samples in Urdu, English, German, and Spanish. Preprocessing of each text instance was accomplished through normalization and tokenization specific to the size of its language with the use of the XLM-RoBERTa-base tokenizer \cite{conneau2020unsupervisedcrosslingualrepresentationlearning}, with all sequences being either truncated or padded to the maximum length of 128 tokens. The division of the dataset into training, validation, and test subsets consists of 70\%, 15\%, and 15\%, respectively, retaining a balanced distribution of classes for a robust evaluation process.

Displayed in Figure~\ref{fig:GHaLIB}, the pipeline begins by identifying the language of the input and assigning it to a proper contextual encoder: UrduBERT \cite{khalid2021rubertbilingualromanurdu} for Urdu, RoBERTa \cite{liu2019robertarobustlyoptimizedbert} for English, and EuroBERT \cite{boizard2025eurobertscalingmultilingualencoders} for German and Spanish. These encoders generate dense embeddings, which are subsequently concatenated and fed into a context-based classifier using XLM-RoBERTa-base \cite{conneau2020unsupervisedcrosslingualrepresentationlearning} as the pretrained transformer. Models were also tested on larger variants (e.g., XLM-RoBERTa-large) through comparative experiments, which showed higher accuracy and improved performance at a justified cost. However, the base version was chosen due to its superior balance of cost-efficiency and feasibility for training on low-end Graphics Processing Units.

To handle the class imbalance, in particular the risk for false negatives, the custom WeightedTrainer \cite{chen2022weightedtrainingcrosstasklearning} comes with a cross-entropy loss where the weight is 1.5× for the positive (hope) class. Hyperparameter tuning is done by means of Optuna \cite{akiba2019optuna} over 30 trials, with a search among the given learning rates (5e-6 to 5e-5), batch sizes (4, 8, 16), warm-up ratio (0.0 to 0.3), weight decay (0.0 to 0.1), and dropout rates (0.1 to 0.3).

After training, the model classification threshold, ranging between 0.3 and 0.8, is tuned on the validation set to optimally balance the trade-off between false negatives and false positives. The evaluation metrics include accuracy, weighted F1-score, and confusion matrices, with explicit consideration of FN and FP rates. All visualizations of curves and matrices are presented in high-quality resolution for clarity.

All these experiments were conducted in Python (v3.12+) using Hugging Face Transformers on the Kaggle environment with 2×16GB NVIDIA T4 GPUs. For full reproducibility, all code, preprocessing scripts, and configuration files are versioned and made publicly available in a dedicated \href{https://github.com/ahmedembeddedxx/GHaLIB/}{repository}. Random seeds were fixed across NumPy and PyTorch to reduce variability in the results. Data splits (train/validation) are stratified and generated via deterministic shuffling so that the partitions remain equivalent for all experiments, and the final reported results were computed on hidden benchmark of CodaLab. For every run of the experiment, we preserved and documented model checkpoints as well as hyperparameter settings. Performance metrics and visualizations were created with a standard script and printed at 300 DPI in Times New Roman font for publication clarity. The use of Optuna \cite{akiba2019optuna} for hyperparameter optimization is logged with fixed random states to allow complete reruns of the search process.

This ensures that our study enables independent verification, benchmarking, and future experimentation under enhanced conditions.

\section{Results \& Experimentation} \label{sec:results}

A comprehensive evaluation was conducted across multiple transformer-based architectures for both binary and multi-class hope speech classification in Urdu, English, German, and Spanish. The tested models included XLM-RoBERTa with language-specific encoders such as UrduBERT, EuroBERT, mBERT \cite{devlin2019bertpretrainingdeepbidirectional}, as well as general-purpose encoders such as RoBERTa 
\cite{liu2019robertarobustlyoptimizedbert}, DistilBERT \cite{sanh2020distilbertdistilledversionbert}, MultiLing BERT \cite{devlin2019bertpretrainingdeepbidirectional}, and GloVe \cite{Brochier_2019}. Performance was assessed using accuracy, recall, precision, and F1 (macro) score. Full experimental results are summarized in Table~\ref{tab:metrics}, with abbreviations in Table~\ref{tab:acronyms}.

\begin{table}[!t]
    \caption{List of acronyms used in this metrics table.}
    \label{tab:acronyms}
    \centering
    \begin{tabular}{ll}
        \toprule
        \textbf{Acronym} & \textbf{Full Form} \\
        \midrule
        ADB & AdaBoost Classifier \\
        SVM & Support Vector Machine \\
        LGB & LightGBM Classifier \\
        Multi & Multi-class Classification \\
        Binary & Binary Classification \\
        \bottomrule
    \end{tabular}
\end{table}

\begin{table*}[!t]
    \caption{Performance metrics}
    \label{tab:metrics}
    \centering
    \begin{tabular}{lcccccccc}
        \textbf{Architecture} & \textbf{Model} & \textbf{Type} & \textbf{Language} & \textbf{Accuracy} & \textbf{Recall} & \textbf{Precision} & \textbf{F1 Macro} & \textbf{Ranking}\\
        \hline
         
         & & & & & &\\
         
        XLM RoBERTa + UrduBERT & - & Binary & Urdu & 95.0\% & 95.0\% & 95.0\% & \greenbf{95.0\%} & 1st \\
        RoBERTa Urdu & ADB / SVM / LGB & Binary & Urdu & 94.8\% & 95.0\% & 94.8\% & 94.8\% & \\
        UrduBERT & ADB / SVM / LGB & Binary & Urdu & 94.6\% & 94.8\% & 94.8\% & 94.6\% & \\
        RoBERTa Urdu & ADB / SVM / LGB & Binary & Urdu & 93.6\% & 93.6\% & 93.8\% & 93.6\% & \\
        MultiLing BERT & ADB / SVM / LGB & Binary & Urdu & 93.6\% & 93.6\% & 93.8\% & 93.6\% & \\
         & & & & & & \\
        
        RoBERTa Urdu & ADB / SVM / LGB & Multi & Urdu & 78.3\% & 65.6\% & 65.4\% & \greenbf{65.2\%} & 1st \\
        MultiLing BERT & ADB / SVM / LGB & Multi & Urdu & 78.0\% & 65.2\% & 64.3\% & 64.2\% & \\
        XLM RoBERTa + UrduBERT & - & Multi & Urdu & 77.1\% & 62.6\% & 62.8\% & 62.5\% & \\
        UrduBERT & ADB / SVM / LGB & Multi & Urdu & 76.7\% & 62.9\% & 62.2\% & 62.3\% & \\
        RoBERTa + ROS & ADB / SVM / LGB & Multi & Urdu & 74.4\% & 54.3\% & 70.3\% & 52.6\% & \\
         & & & & & & \\
        
        XLM RoBERTa & - & Binary & English & 86.3\% & 86.3\% & 86.3\% & \greenbf{86.3\%} & 2nd \\
        Finetuned RoBERTa & - & Binary & English & 85.2\% & 85.3\% & 85.2\% & 85.2\% & \\
        RoBERTa & ADB / SVM / LGB & Binary & English & 85.2\% & 85.3\% & 85.2\% & 85.2\% & \\
        GloVe & ADB / SVM / LGB & Binary & English & 84.3\% & 84.3\% & 84.4\% & 84.3\% & \\
        MultiLing BERT & ADB / SVM / LGB & Binary & English & 76.0\% & 76.0\% & 76.0\% & 76.0\% & \\
        DistilBERT & ADB / SVM / LGB & Binary & English & 67.0\% & 54.2\% & 54.2\% & 54.0\% & \\
         & & & & & & \\
        
        XLM RoBERTa & - & Multi & English & 77.2\% & 72.3\% & 70.3\% & \greenbf{71.0\%} & 2nd \\
        DistilBERT & ADB / SVM / LGB & Multi & English & 77.0\% & 70.2\% & 72.0\% & 71.0\% & \\
        Finetuned RoBERTa & - & Multi & English & 76.2\% & 70.2\% & 70.5\% & 70.2\% & \\
        RoBERTa & ADB / SVM / LGB & Multi & English & 76.7\% & 69.7\% & 78.8\% & 70.3\% & \\
        MultiLing BERT & ADB / SVM / LGB & Multi & English & 74.7\% & 68.7\% & 70.7\% & 69.2\% & \\
        GloVe & ADB / SVM / LGB & Multi & English & 70.3\% & 62.7\% & 62.8\% & 62.7\% & \\
         & & & & & & \\
        
        XLM RoBERTa + EuroBERT & - & Binary & Spanish & 85.0\% & 85.1\% & 85.0\% & \greenbf{85.0\%} & 1st \\
        XLM RoBERTa + mBERT & - & Binary & Spanish & 84.5\% & 84.5\% & 84.5\% & 84.5\% & \\
         & & & & & & \\
        
        XLM RoBERTa + EuroBERT & - & Multi & Spanish & 75.3\% & 68.5\% & 69.6\% & \greenbf{68.5\%} & 3rd \\
        XLM RoBERTa + mBERT & - & Multi & Spanish & 74.0\% & 67.0\% & 70.0\% & 67.7\% & \\
         & & & & & & \\
        
        XLM RoBERTa + EuroBERT & - & Binary & German & 87.1\% & 87.1\% & 87.4\% & \greenbf{87.4\%} & 2nd \\
        XLM RoBERTa + mBERT & - & Binary & German & 86.0\% & 86.4\% & 86.6\% & 87.0\% & \\
        Finetuned RoBERTa & - & Binary & German & 86.7\% & 86.4\% & 86.4\% & 86.4\% & \\
         & & & & & & \\
        
        XLM RoBERTa + EuroBERT & - & Multi & German & 83.4\% & 70.9\% & 70.5\% & \greenbf{70.1\%} & 1st \\
        XLM RoBERTa + mBERT & - & Multi & German & 83.0\% & 73.5\% & 66.9\% & 67.7\% & \\
    \end{tabular}
\end{table*}

In the binary classification experiments, XLM-RoBERTa-Base obtained a 95.0\% accuracy in Urdu, the highest among all languages and settings, followed by 87.4\% in German, 86.3\% in English, and 85.0\% in Spanish. The results thus show its ability to generalize with languages differing in scripts and structures. This suggests that multilingual pretraining allows the model to effectively share representations across linguistically diverse inputs, improving cross-lingual transfer. In multi-class classification, the model performed best with the performance results of 71.0\% in English, 70.1\% in German, 68.5\% in Spanish, and 65.2\% in Urdu. These differences probably arise from the fact that the amount of English and European languages annotated datasets is much larger than that of Urdu, which allows for a better learning of representations in the context. The results of the model continued to show excellent performance despite the extra complication like overlapping words and larger context in multi-class, thus confirming its adaptability and robustness. 

Language-specific models were found to be the most benefit for morphologically rich and lower-resource languages, because they can capture morphological variations and syntax structures that multilingual models often overlook, particularly Urdu and German, where the difference from generic multilingual models was most striking. We also found that general-purpose transformer models performed well in English, likely due to abundant high-quality training corpora and the dominance of English in model pretraining, while models without clear language adaptation, such as DistilBERT and GloVe, consistently underperformed. This situation could be explained by the fact that there is a lot of high-quality pretraining data and resources available for English, which lets the models that are trained for general purposes to detect the subtle contextual patterns more efficiently than low-resource languages. Recall and precision tracked overall F1-scores neatly for most languages, but the modest precision gains suggested that misclassifications arose from difficult or ambiguous cases rather than from systematic biases. This pattern indicates that the model’s errors were balanced rather than biased toward false positives or negatives, suggesting stable decision thresholds.

Tuning thresholds on the validation data yielded small gains in the balance of false negatives to false positives, while producing no overall change in model rankings. Based on error analysis, even skilled teams found consistent challenges with code-mixed and contextually ambiguous inputs, especially in the case of multi-class tasks. In sum, findings indicate that integrating language-specific encoders on top of the multilingual transformer backbone yields state-of-the-art results in hope speech detection, particularly in binary classification and, more importantly, in low-resource languages. These findings highlight that beyond model architecture, linguistic ambiguity and code-mixing remain key challenges that limit perfect classification performance.

\section{Conclusion \& Future Work} \label{sec:future}
The findings provided by GHaLIB shows the strength of XLM-RoBERTa-Base as a trustworthy and strong contender for multilingual text classification. Its consistent cross-domain effectiveness makes it a reliable option for any real-world multilingual NLP applications, especially when considering computational resources. Future work can further enhance the framework by exploring experimentation on more low-resourced languages such as Punjabi, Seraiki, and Sindhi, which are heavily spoken languages in Pakistan, along with other such regional languages. This will add to the multilingual scope of hope speech classification and contribute to a more inclusive NLP resource by extending such languages into the evaluation pipeline. In addition, techniques like domain-specific pretraining and parameter-efficient fine-tuning, can further strengthen the adaptability and scalability of this framework in real-world applications.

\bibliographystyle{plain} 
\bibliography{references}  
\end{document}